%% file: neurips_2020.tex
\documentclass{article}

    \usepackage[preprint]{neurips_2020}

\usepackage[utf8]{inputenc} 
\usepackage[T1]{fontenc}    
\usepackage{hyperref}       
\usepackage{url}            
\usepackage{booktabs, caption}       
\usepackage{amsfonts}       
\usepackage{nicefrac}       
\usepackage{microtype}      
\usepackage{color}
\usepackage{amsmath}
\usepackage{algorithm2e}

\usepackage{graphicx}
\usepackage[toc,page]{appendix}
\usepackage{multirow} 
\usepackage{makecell}
\usepackage{subcaption}
\usepackage{wrapfig}
\usepackage{tablefootnote}
\usepackage{color}
\usepackage{float}

\usepackage{enumitem,kantlipsum}

\title{Robustness to Adversarial Attacks in Learning-Enabled Controllers}

\author{%
  Zikang Xiong 
    \\
  Purdue University\\
  \texttt{xiong84@purdue.edu} \\
   \And
   Joe Eappen \\
   Purdue University \\
   \texttt{jeappen@purdue.edu} \\
   \AND
   He Zhu \\
   Rutgers University \\
   \texttt{he.zhu.cs@rutgers.edu} \\
   \And
   Suresh Jagannathan \\
   Purdue University \\
   \texttt{suresh@cs.purdue.edu} 
}

\begin{document}

\maketitle

\begin{abstract}
  Learning-enabled controllers used in cyber-physical systems (CPS) are
  known to be susceptible to adversarial attacks.  
  Such attacks manifest as perturbations to the states generated by the
  controller's environment in response to its actions.  
  We consider state perturbations that encompass a wide variety of adversarial attacks and describe an attack scheme for discovering adversarial states. 
  To be useful, these attacks need to be \emph{natural}, yielding states in which
  the controller can be reasonably expected to generate a meaningful
  response.  We consider shield-based defenses as a means to improve
  controller robustness in the face of such perturbations.  Our
  defense strategy allows us to treat the controller and environment
  as black-boxes with unknown dynamics. We provide a two-stage
  approach to construct this defense and show its effectiveness
  through a range of experiments on realistic continuous control
  domains such as the navigation control-loop of an F16 aircraft and
  the motion control system of humanoid robots.
\end{abstract}

\setcitestyle{square,numbers} 
\section{Introduction}

Deep reinforcement learning (RL) approaches have shown promise in
synthesizing high-quality controllers for sophisticated cyber-physical
domains such as autonomous vehicles and robotics~\citep{BP+18,DR+17}.
However, because these domains are characterized by complex
environments with large feature spaces, they have proven vulnerable to
various kinds of adversarial attacks~\citep{GD+20} that trigger
violations of safety conditions the controller must respect.  For
example, a safe controller for an unmanned aerial navigation system
must account for a myriad of factors with respect to
weather, topography, obstacles, etc., that may maliciously
affect safe operation, only a fraction of which are likely to have been 
considered during training.  Ensuring that controllers are robust in the face 
of adversarial attacks is therefore an important ongoing challenge.

Although the state space over which the controller's actions take
place is intractably large, the need to conform to physical realism
greatly reduces the attack surface available to an adversary, in
practice.  For example, an adversarial attack on a UAV cannot generate
an obstacle from thin air, suspend gravity, or instantaneously
eliminate prevailing wind conditions.  A meaningful attack is thus
expected to generate states that are \emph{natural}~\citep{GD+20},
i.e., states that can be derived from perturbations of
realizable states as defined by the physics of the application domain
under consideration.  We are interested in identifying such states
along a trajectory representing a rollout of the controller's learnt
policy.

Successful attacks manifest as a safety failure in this policy.  While
these kinds of falsification methods are certainly
important~\citep{ghosh_verifying_2018} to assess the overall safety of
a controller, devising a defensive strategy that prevents unsafe
operation resulting from such attacks is equally valuable.  In
black-box, model-free RL settings where a policy's internal structure
is unknown, shield-based defenses are typically
required~\citep{alshiekh_safe_2017}.

This paper considers the generation of such defenses that comprise
three distinct components: (1) a mechanism to efficiently explore
adversarial attacks within a realistic bounded area of the states
found in simulated and seemingly safe trajectories; (2) a detector
policy that predicts safe and unsafe states based on previously
observed successful attacks; and, (3) an auxiliary policy that is
triggered whenever the detector identifies an unsafe state; the role
of the policy is to propose a new action that re-establishes a safe
trajectory.  To be practical, each of these components requires new
insights on adversarial attack and defense in RL.  Effective attack
and detection strategies in realistic cyber-physical environments must
necessarily reduce a high-dimensional state space to a lower
dimensional one focused on features most relevant to an adversarial
attack.  While an effective defense must be tuned with respect to the
detector to protect against identified attacks, any useful methodology
must also be able to generalize the shield to successfully prevent
unseen attacks, i.e., attacks for which the detector has not been
trained against.

\begin{figure}
  \centering
  \includegraphics[width=\textwidth,scale=.5]{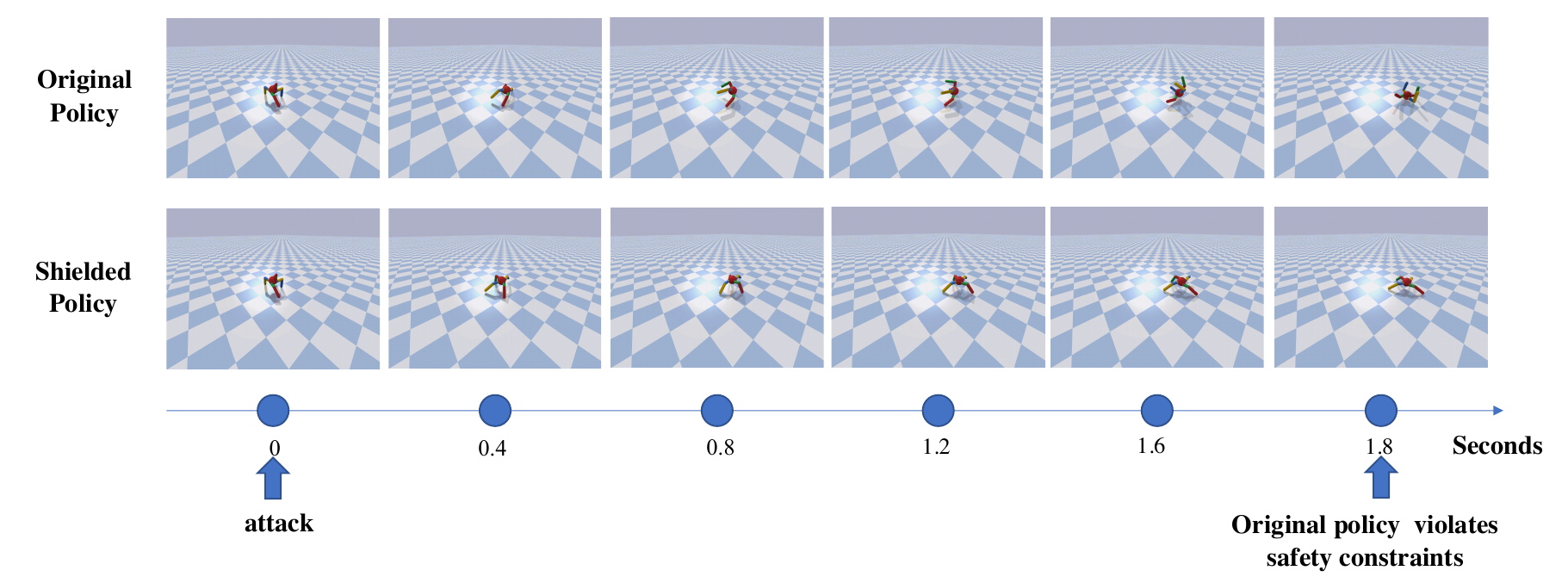}
  \caption{\small Simulation of a shielding defense against an
    adversarial attack on an Ant robotics controller.  The top row
    shows the effect of a successful attack.  The bottom row
    illustrates an execution starting from the same initial (unsafe)
    state augmented with a shielded defense.}
\end{figure}

Our contributions are three-fold:

\begin{enumerate}

\item We present an adversarial testing framework that identifies
  natural unsafe states found near safe ones in a simulated trajectory
  of learning-enabled controllers.  The RL controller is guaranteed to
  produce trajectories that lead to a safety violation from these
  states.

\item We learn detector classifier from these unsafe trajectories that
  precisely classify safe and unsafe states, and auxiliary
  policies triggered when the detector identifies an unsafe state;
  applying a shield on a presumed unsafe state yields a new safe
  trajectory, i.e., a trajectory that satisfies desired safety
  properties.  

\item We provide a detailed experimental evaluation over a range of
  well-studied RL benchmarks, including the PyBullet
  suite of robotic environments and the F16 ground collision avoidance
  system.  Our results justify our claim that the robustness of
  black-box RL-enabled controllers can be significantly enhanced using
  learning-based detection and shielding approaches, without noticeable
  loss in performance.

\end{enumerate}

Our overarching goal in this paper is to overcome inherent
deficiencies in RL policies that do not account for environment
conditions which can be maliciously or unintentionally perturbed.
These conditions potentially comprise a large attack surface, but in
the black-box setting we consider here, fine-tuning the model is not
possible.  Our approach thus provides a meaningful pathway to identify
and mitigate attacks that exploit unrecognized weaknesses of the
deployed model, resulting in learning-enabled RL controllers that
provide much higher degrees of assurance than the state-of-the-art,
\emph{without} requiring re-training or white-box analysis.

The remainder of the paper is organized as follows.  In the next
section, we present some relevant background material.
Section~\ref{sec:statement} formalizes the problem we consider and our
goals.  Section~\ref{sec:approach} presents details about our
approach.  Experimental results are given in~\ref{sec:expt}. Related
work and conclusions are presented in Sections~\ref{sec:related}
and~\ref{sec:conc}.

\section{Background}
\label{sec:background}

\paragraph{Optimum Policies for Discrete-time Continuous Systems}
We consider discrete-time systems $M=\langle S,A,f,R \rangle$ with non-linear dynamics $f$, continuous state space $S\subseteq \mathbb{R}^n$ and action space $A\subseteq \mathbb{R}^k$.
Here $s_{t+1} = f(s_t,u_t)$ where $s_t\in S$ is the state at time step $t$ and $f$ is a complex unknown non-linear function. The objective, given a reward function  $R:S\times A \rightarrow \mathbb{R}$, is to find a policy $\pi: S\rightarrow \Delta(A)$ where $\Delta(A)$ is a probability distribution over $A$  that maximizes the discounted return $\mathbb{E}_{\pi,S_o}[\underset{t=0..T}{\sum} \gamma^t R(s_t,a_t)]$ where $\gamma\in(0,1)$ is the discount factor and $S_o$ is the initial state space. Given a policy $\pi$ a trajectory $\tau=(s_0,...,s_T)$ is the sequence of states obtained when $s_{t+1} = f(s_t,\pi(s_t))$ up to a horizon $T$.


\paragraph{Bayesian Optimization}
Bayesian optimization (BO) is a black-box optimization approach which aims at finding the global minimum for a black-box function $F(x)$ by querying a constrained number of times. This constraint is motivated by the expectation that evaluating the objective function often requires significant computational resources. 
BO models $F(x)$ with a prior (frequently Gaussian Process-based) and uses an acquisition function to query values of $F(x)$ while trading off exploration and exploitation.

\section{Problem Statement} 
\label{sec:statement}

\newcommand{\attack}{\texttt{Att}_\epsilon}
\newcommand{\advset}{\texttt{Adv}_\epsilon}


Consider a continuous-state system $M$ where user-defined \emph{safe} ($S_s$) and \emph{unsafe} states ($S_u$) form a partition over the entire state space $S$ of $M$.
A rollout of a trained agent policy $\pi_o$ on $M$ yields a state trajectory $\tau_o = (s_0,...,s_T)$. 
Let $\mathcal{T}(\pi_o,M,S_o)$ be the set of \emph{safe} trajectories formed by rollouts of policy $\pi_o$ in the system $M$ with initial state space $S_o$
(\textit{i.e.} trajectories in this set do not contain any states in $S_u$).

We consider the state dimensions which are used for computing the input of the agent policy $\pi_o$. These are often related to spatial characteristics such as joint position, angle, velocity, and robot center of mass. Define a filter $\phi(s)$ that yields a subset of modifiable state dimensions. Applying $\phi(s)$ to a trajectory $\tau$ yields $\phi(\tau) = (\phi(s_0),...,\phi(s_T))$.

Define a $\epsilon$-state perturbation of a state $s$ on the filtered dimensions in $\phi(s)$ to be another state $s'\in S$ that is within $\epsilon$ distance away from $s$ by varying the set of dimensions in $\phi(s)$. We choose the $L_\infty$ norm to measure state distances. Given a state $s$, denote $P_\epsilon(s)$ to be the set of all possible perturbations on $s$. 

\begin{wrapfigure}{H!}{0.39\textwidth}
    \begin{minipage}{0.45\textwidth}
      \begin{algorithm}[H]
  \DontPrintSemicolon
  \SetKwFunction{Shield}{Shield}
  \SetKwProg{Fn}{Function}{:}{}
  \Fn{\Shield{$state$}}{
  Given $\texttt{detector}$,  $\pi_o$, $\pi_{\mathit{aux}}$
  
        \eIf {\texttt{detector} ($state$)}{   
            \KwRet $\pi_{\mathit{aux}}$\; 
        }{
            \KwRet $\pi_o$\;
        }
  }
  \label{algo:def}
  \caption{Shield Defense Outline}
\end{algorithm}
    \end{minipage}
  \end{wrapfigure}
We define an \emph{Adversarial State Attack} $\attack(\tau_o)$ to be a set of states in which each state $s \in \attack(\tau_o)$ is an $\epsilon$-state perturbation on a state in the trajectory $\tau_o$ that causes the new trajectory to be unsafe. 
In other words, $s \in \attack(\tau_o)$ if, for some $k$ and $s'$, $0\leq k \leq T$, $s'\in P_\epsilon(s_k)$, and executing the policy $\pi_o$ from $s'$ leads to a state in $S_u$. This can be thought of as a state perturbation at an arbitrary time index along the trajectory that results in safety failure. 
These perturbations can be minute, a small shift of a robot due to a bump on the ground, or mild turbulence affecting the positioning of a jet in midair. 
If $\epsilon$ is unconstrained, perturbations can be unrealistic, unlikely to occur in practice, or impossible to defend against given existing environment transition dynamics. 
We thus concern ourselves with values of $\epsilon$ that lead to perturbed states in Section \ref{sec:epsilon_selection}.

We aim to find a way to defend against adversarial state attacks leading to a more robust control strategy that is still performant. 
In real-world black-box scenarios, we have no access to both controller internals (underlying control policy) and environment conditions.
Thus, an effective defense must treat the original policy $\pi_o$ as a black box and make no assumptions about the (hidden) environment dynamics. 
With these constraints in mind, we divide this problem into two stages. The first step is to detect adversarial state attacks, knowledge of which can be used to act differently, by
shielding $\pi_o$ from malicious state perturbations (the second step).
One defense outline is in the style of \citep{alshiekh_safe_2017} where a shield modulates actions taken by $\pi_o$ based on the observation provided by the environment. We structure a simple two-level function to represent the shield shown in Algorithm~\ref{algo:def}. 
The upper level would be able to detect that an $\attack$ attack has occurred while the lower level would be a policy to bring the system back to a safe state. 
We call the upper level function the \textit{detector} 
and the lower level policy the \textit{auxiliary policy} ($\pi_{\mathit{aux}}$).

With this setup, the constructed defense then uses the detector to decide whether to follow the original policy $\pi_o$ or switch to $\pi_{\mathit{aux}}$.
If these two functions are accurate, this approach is robust to adversarial state attacks for a given $\epsilon$.
Our goal is thus to learn these two functions keeping in mind the restrictions placed on the defense scheme.



\section{Approach}
\label{sec:approach}


Based on the definition of Adversarial State Attack $\attack$ (Section~\ref{sec:statement}), given a trained policy $\pi_o$, a system $M$ with some initial states $S_o$, we aim to find a set of state-based adversarial attacks $\advset(\pi_o, M, S_o)$ defined as:
$$\advset(\pi_o, M, S_o) = \{s\ \vert\ {s \in \attack(\tau)} \wedge \tau \in \mathcal{T}(\pi_o,M,S_o)\}$$ 
Our defense strategy follows the outline described in Algorithm~\ref{algo:def}.

\subsection{Safety Specifications}

Similar to~\cite{ghosh_verifying_2018}, we consider a safety specification $\varphi$ consisting of multiple predicates denoted as $\rho$ connected using Boolean operators including conjunction and disjunction:
$$
\varphi := \rho\ \vert\ \varphi \wedge \varphi\ \vert\ \varphi \vee \varphi\ \ \ \ \rho := t = t'\ \vert\ t \neq t'\ \vert\ t \le t'\ \vert\ t<t'\ \vert\ t \ge t'\ \vert\ t>t'
$$
A term $t$ is any real-valued function defined over system variables. For example, consider a robot with the height of the center of mass being $z$.  One safe height specification $\varphi_{\mathit{safe}}$ is $z_{\mathit{safe}} < z$ where $z_{\mathit{safe}}$ is a height threshold.

\subsection{Attack}
\label{sec:att_approach}

The search problem for $\advset(\pi_o, M, S_o)$ as defined above is in general intractable. We thus require a more precise objective that is easy for optimization.
To tractably find $\advset(\pi_o, M, S_o)$, we define a \textit{safety reward} function $L(\varphi):S \rightarrow \mathbb{R}$ for $M$. A key feature of $L(\varphi)$ is that, given a state $s \in S$, $\varphi$ holds on $s$ iff $L(\varphi)(s) > 0$. We define $L(\rho)$ recursively. First, $L(t<t') := t'- t$. For the above robot example, $L(\varphi_{safe}) := z-z_{safe}$. We have $L(t = t') := \delta[t = t']$ where $[\cdot]$ is an indicator function and $\delta$ is a user-configurable constant, $L(t \neq t') := L(t < t' \vee t > t')$, and $L(t \le t') := L(t < t' \vee t = t')$. $L(\varphi)$ is also recursively defined: $L(\varphi \wedge \varphi') := \min (L(\varphi), L(\varphi'))$ and $L(\varphi \vee \varphi') := \max(L(\varphi), L(\varphi'))$. 

Given a trajectory $\tau$, its \textit{safety reward} is
\begin{align*}
L(\varphi)(\tau)=\underset{s\in \tau}{\min}\ L(\varphi)(s)
\end{align*}




\paragraph{Attack setup} 


With $L(\varphi)$ defined above as an effective proxy for safety measurement, given a (set of) trajectory $\tau$ of a system $M$ from $S_o$ i.e. $\tau \in \mathcal{T}(\pi_o,M,S_o)$, we search in $\underset{s \in \tau}{\bigcup}{P_\epsilon(s)}$ a set of state $S'$ with the objective of minimizing the safety reward $L(\varphi)(\tau')$ where $\tau'\in\mathcal{T}(\pi_o,M,S')$.
Observe that $L(\varphi)(\tau')$ is essentially a function over states (parameterized by the first state of $\tau'$ and the policy $\pi_o$).
The ability of Bayesian Optimization (BO) in optimizing black-box functions (e.g. $L(\varphi)(\tau')$) makes it qualified as an optimization scheme for our purpose. 
We thus use BO to find states in $\advset(\pi_o, M, S_o)$ that lead to trajectories with a negative $L(\varphi)$ reward and hence unsafe. 
In our experiments, we use EI (Expected Improvement) \citep{jones1998efficient} as the acquisition function and GP (Gaussian Processes) \citep{rasmussen2003gaussian} as the surrogate model. 
  
  

\paragraph{Feature Selection} 
Na\"ive BO does not scale with the dimensionality of a controller's feature space. One way to improve scalability is to optimize over a reduced set of features. For example, \citet{ghosh_verifying_2018} used REMBO \citep{wang2013bayesian} to create a reduced input space. We apply feature selection using a 2-stage approach. First, we run BO attacks 
without feature selection and collect unsafe and safe trajectories.  We then train a random forest with the collected data, selecting the top-$k$ features based
on a mean decrease in impurity i.e. refining the filter operator $\phi$ in Sec.~\ref{sec:statement}. We rerun the attack on these features again, searching for more unsafe trajectories. 
  
\subsection{Defense}
\paragraph{Training the Detector} Our defense approach must be aware of potential unsafe states that occur when using the original policy $\pi_o$. 
To this end, we train a classifier (the detector) with data derived during the attack phase. When we attack policy $\pi_o$ with BO, we obtain batches of trajectories via simulation in the environment, some of which starting from states in $\advset(\pi_o, M, S_o)$ are unsafe, and others are safe.
For a safe trajectory, we label all included states as safe, while 
for an unsafe trajectory $\tau$, all states that follow a state found in $\attack(\tau)$ are labelled as unsafe. Training a classifier with this data yields a detector that differentiates between states that are safe from those that produce unsafe trajectories affected by states in $\advset(\pi_o, M, S_o)$.

\paragraph{Auxiliary Policy} 
When running a system, we use the detector to monitor current system states. 
If the detector identifies a potentially vulnerable state, the defense strategy must yield an alternative action (sequence) to prevent the agent from entering an unsafe region. The auxiliary policy should be able to defend against attacks in $\attack$.
To obtain the final building block of our defense, we use the trained detector to provide a reward signal for training an auxiliary policy ($\pi_{\mathit{aux}}$)
with PPO \cite{DBLP:journals/corr/SchulmanWDRK17} and DDPG \citep{pmlr-v32-silver14}.

Specifically, we train $\pi_{\mathit{aux}}$ on a slightly modified version of $M$, $M'=\langle S,A,f,R_{M'}\rangle$. The reward function $R_{M'}$ has two components:
\begin{enumerate}[leftmargin=*]
    \item 
We integrate the detector as a part of the reward function to train the auxiliary policy. We choose a simple reward signal based on the detector output. For an input state of the reward function, if the detector finds the state 
safe, the action is given a reward 1 and otherwise a reward 0 is given. We call it \textit{detector reward}. 
This reward component aims to bring the agent into a safe state identified by the detector.
\item The safety reward defined in Sec. \ref{sec:att_approach} is also an indicator of safety. A straightforward way to use this information is to add it as well to the reward signal for training the auxiliary policy. Optimizing the safety reward of each rollout maximizes policy safety. This reward component encourages $\pi_{\mathit{aux}}$ to behave safely.
\end{enumerate}

%
%

\paragraph{Deployment of the defense}
Intuitively, $\pi_{\mathit{aux}}$ only focuses on safety and does not care about performance.
Running the original policy $\pi_o$ in tandem with $\pi_{\mathit{aux}}$ (Algorithm~\ref{algo:def}) 
can retain performance provided by $\pi_o$ while ensuring safety provided by $\pi_{\mathit{aux}}$.

\section{Experiments}
\label{sec:expt}

\paragraph{PyBullet} A number of our experiments are conducted in PyBullet \citep{coumans2019}, a well-studied open-source suite of robotic environments.  
The specific environments tested on were the Ant, HalfCheetah, Hopper and Humanoid robots. The objective of each of these environments is to stay upright and sustain forward motion. 
The robots are modeled using the velocity, orientation, and position of their joints. 
For example,  Ant's 8-D action space is used to control each individual motor while navigating its 29-D state space. Its observation has 28 dimensions consisting of 4 dimensions of feet contact information, and 24 dimensions of joint information with body location and velocity.  Its primary safety constraint is to remain upright; additional details are provided in Appendix A.3. 
For more details of the environments, one may refer to the models' XML definition in the PyBullet repo \footnote{\href{https://github.com/bulletphysics/bullet3/tree/master/examples/pybullet/gym/pybullet_data/mjcf}{PyBullet Mujoco XML definitions}}. 
For the attacked policy, we used an open-source implementation of popular RL algorithms \citep{stable-baselines} with the saved models provided\footnote{\href{https://github.com/araffin/rl-baselines-zoo/tree/625646308d98d92400470e2ff3b5bf8e8913c433}{RL pretrained policy}} which are trained on stochastic versions of the environment (to simulate sensor noise).

\paragraph{F16 Ground Collision Avoid System} The F16 environment is a model of the jet's navigation control system\citep{heidlauf2018verification}.
The F16 is modeled with 16 variables 
and with non-linear differential equations as dynamics. The safety constraints are provided based on the aircraft flight limits and boundaries of the model. 
Our objective is to keep the jet level and flying within the specified constraints.
Further details on the state dimensions, initialization and safety bounds are given in Appendix A.2. 
The attacked policy of this model is trained with PPO.

\paragraph{Classic Control Environments} In addition to the above environments, we include several classical control benchmarks including (Inverted) Pendulum, $n$-Car platoon and large helicopter. 
The (Inverted) Pendulum's goal is to swing a pendulum to vertical. 
$n$-Car platoon models multiple ($n$) vehicles forming a platoon maintaining a safe distance relative to one another \citep{schurmann2017optimal}. 
Large helicopter \citep{gopalakrishnan_spaceex_2011} models a helicopter with 28 variables and has constraints on each variables. These benchmarks' constraints are provided in Appendix A.1.
We trained the Helicopter with DDPG while the other models use the pretrained DDPG \citep{pmlr-v32-silver14} policies given in \citep{zhu_inductive_2019}. 

\subsection{Evaluation}
\begin{table}[!htp]\centering
\setlength{\leftskip}{-5pt}
\scriptsize
\begin{tabular}{lcccccccc}\toprule
{\sf Benchmarks} &{\sf \thead{dims \\(obs/state)}}& {\sf \thead{simu. traj. \\ length}} &{\sf \thead{Attack \\ $\varepsilon$}} & {\sf \thead{rand. \\ attack}} &{\sf \thead{BO \\ attack}} &{\sf \thead{defense \\succ. rate}} &{\sf \thead{\tiny attack improvement \\ \tiny using shielded policy}} \\\midrule
Hopper-a2c &\multirow{3}{*}{15/12} &\multirow{3}{*}{1000} &\multirow{3}{*}{0.001} &1.29\% &2.18\% &91.40\% &43.51\% \\
Hopper-ppo & & & &0.41\% &5.27\% &97.80\% &27.97\% \\
Hopper-trpo & & & &1.71\% &1.97\% &97.90\% &48.16\% \\\midrule
HalfCheetah-a2c &\multirow{7}{*}{26/17} &\multirow{7}{*}{1000} &\multirow{7}{*}{0.02} &0.36\% &9.43\% &91.40\% &34.46\% \\
HalfCheetah-acktr & & & &0.90\% &14.54\% &92.80\% &65.23\% \\
HalfCheetah-ddpg & & & &1.63\% &11.84\% &88.20\% &67.27\% \\
HalfCheetah-ppo2 & & & &0.38\% &4.68\% &97.80\% &50.53\% \\
HalfCheetah-sac & & & &4.40\% &5.88\% &97.90\% &62.12\% \\
HalfCheetah-trpo & & & &0.77\% &4.83\% &97.90\% &49.02\% \\\midrule
Ant-a2c &\multirow{6}{*}{28/29} &\multirow{6}{*}{1000} &\multirow{6}{*}{0.0075} &0.17\% &1.26\% &82.60\% & 66.20\% \\
Ant-ddpg & & & &0.42\% &4.43\% &94.30\% &93.57\% \\
Ant-ppo & & & &1.47\% &11.78\% &88.70\% &91.39\% \\
Ant-sac & & & &2.82\% &14.83\% &96.00\% &94.02\% \\
Ant-td3 & & & &5.79\% &12.79\% &93.60\% &72.52\% \\ \midrule
Humanoid-ppo &44/47 &1000 &0.001 &0.38\% &2.00\% &91.90\% &22.93\% \\\midrule
Pendulum-ddpg &2/2 &200 &* &0.03\% &8.72\% &100.00\% &100.00\% \\
4carploon-ddpg &7/7 &1000 &* &0.01\% &3.82\% &100.00\% &94.71\% \\
8carploon-ddpg &15/15 &2000 &* &0.72\% &4.13\% &100.00\% &100.00\% \\
Helicopter-ddpg &28/28 &2000 &* &0.00\% &0.26\% &100.00\% &100.00\% \\\midrule
F16-ppo &8/16 &2000 &* &0.01\% &2.60\% &96.40\% &96.69\% \\
\bottomrule
\end{tabular}

* Use allowed variance of initial state which is given by the benchmark itself
\caption{Experimental Results. Benchmarks are of the form, E-A where E is the Environment and A is the type of controller attacked.} 
\label{tab: eval_table}
\end{table}

\paragraph{Selecting Attack Range}
\label{sec:epsilon_selection}
As mentioned in Sec.~\ref{sec:statement}, an Adversarial State Attack $\attack$
requires a meaningful $\varepsilon$-state perturbation.
We choose $\varepsilon$ with following strategy. 
For a specific robot in the PyBullet benchmarks, we initialized our $\varepsilon$ to be 0.001. We randomly perturbed initial states and run simulations 1000 times for every available policy. If there is a policy that does not find any unsafe states, we increase $\varepsilon$ by 0.0005, and repeat this process until unsafe states are found in all policies.
We believe this is a sensible strategy because none of these benchmarks
are equipped with constraint specifications that indicate the range of
acceptable states that can be generated by the environment.  For all the other benchmarks,
constraints on these ranges were available and used directly.

\paragraph{Attack Results}
\begin{wrapfigure}{!H}{0.5\linewidth}
    \includegraphics[width=\linewidth]{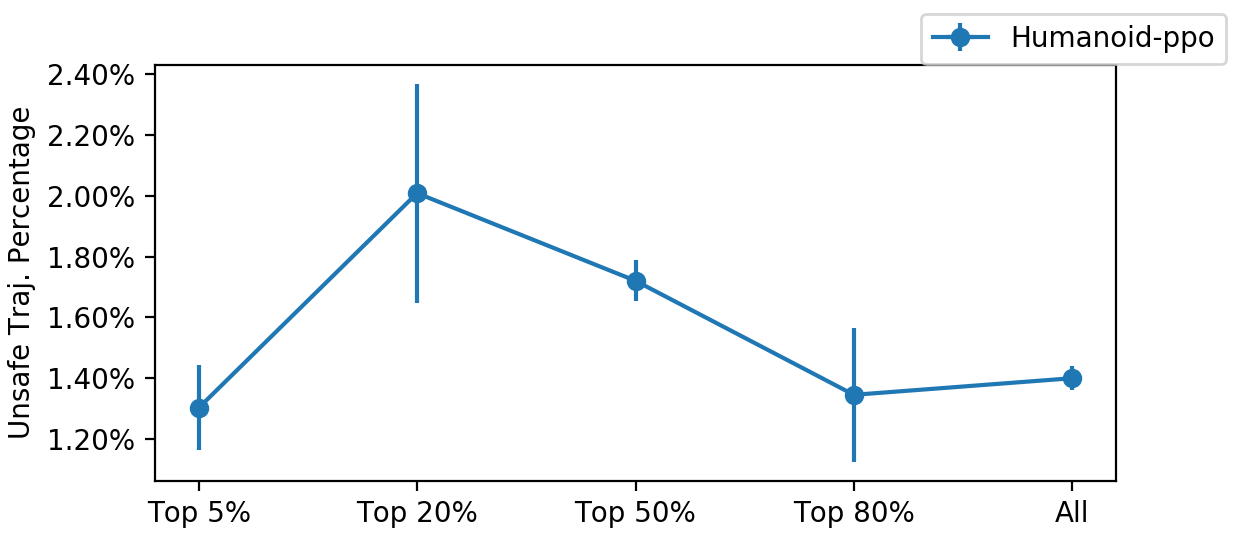}
    \caption{Attack success rate for Humanoid-ppo under different feature dimension reduction percentages.}
    \label{fig: BO_humanoid}
\end{wrapfigure}

Table~\ref{tab: eval_table} shows two attack strategies.  The first,
{\sf rand. attack} indicates the percentage of safety violations
detected when initializing the state randomly around a sampled
trajectory with respect to the chosen $\varepsilon$. For each
benchmark, we randomly sampled 40K states within the bounds determined
by $\varepsilon$ and initialized the simulations accordingly.  Column
{\sf BO attack} shows the attack success rate when using BO to
discover attacks.  The GPs of the method are initialized with 10
randomly sampled states and the BO algorithm samples the safety reward
30 times.  For each state in one trajectory, we generate 40 sampled
trajectories. If the length of an attacked trajectory was $L$, $40
\times L$ trajectories would be collected in one attack. We then count
the unsafe trajectories in these simulations. 
For high-dimensional
benchmarks like Humanoid, our attack focused on the top 20\% most
important features (9 features); these features were selected using
the approach described in \ref{sec:att_approach}.  With this feature
reduction in place, our technique achieves an average 2\% attack
success rate, compared to just a 0.38\% success rate using a randomized
attack strategy.  Results for other dimension reduction choices are
given in Figure~\ref{fig: BO_humanoid}.



\paragraph{Defense Results}
We trained detectors with the states generated by our attack and
learned auxiliary policies with the detector and corresponding safety
reward.  We measured the quality of our defense in two ways.  First,
we want to know how successful our detector and auxiliary policies are
in preventing attacks on the original policy that would otherwise lead
to a safety violation.  Second, given a shielded policy, i.e., a
policy that integrates the original, defense, and auxiliary policy, we
measure how effective this combined policy is in reducing the number
of unsafe states available to an attack, compared to the original
policy.


To answer the first question, we run BO attacks on the original
policy, employing our detector and shield policy to detect and avoid unsafe
trajectories. Note that we may still find unsafe trajectories because
not all attacks are necessarily preventable for reasons discussed
below.  The defense success rate on the PyBullet benchmarks ranges
from 82.6\% to 100\%. On the simpler classical control benchmarks, the
defense success rate can achieve 100\% while the F16GCAS benchmark has
a 96.4\% defense success rate.  These results support our claim that
$\varepsilon$-state perturbations used in adversarial attacks in
these environments can be effectively mitigated.  We note that our
defense method does not provide verifiable guarantees of safety.
There are two primary reasons for this: \textbf{a)} the possible
existence of states in our attack which are not recoverable by any
policy (eg: a robot whose forward momentum will cause it to fall
regardless of any possible shield action that may be taken); and,
\textbf{b)} covariate shift due to our black-box setting that does not
provide insight into environment dynamics, making it impossible to
provide guarantees on detector accuracy during testing since the
detector and auxiliary policy are trained only with the adversarial
states discovered by previously seen BO attacks.  In other words, even
if the detector is accurate, it is also untenable to assert that the
auxiliary policy ($\pi_{\mathit{aux}}$) trained by RL is always
reliable due to variations while training.

The {\sf attack improvement using shielded policy} column shows the
percentage reduction in unsafe states discovered by a BO attack when
applied to the shielded policy as compared to the original.  The
trajectories admitted by the shielded policy are more constrained than
the original since the policy is derived as a mixture of both the
original and the auxiliary (safety) policy.  Consequently, we would
expect a fewer number of unsafe states to be discoverable under this new
policy when compared against a policy in which safety was not taken
into account.  Indeed, the results shown in the column justify this
intuition.  For all benchmarks, the use of a shielded policy reduces
the number of unsafe states found by a BO attack by at least 22.93\%.
The use of this blended policy on the three Classical Control tasks
demonstrates 100\% empirical robustness, while the F16 benchmark is
very close behind at 96.69\%.  Several of the PyBullet benchmarks such
as Ant-ddpg and Ant-sac show similar improvement.  

\begin{figure}
    \centering \setlength{\leftskip}{-120pt}
    \includegraphics[width=1.5\textwidth,
      height=0.28\textwidth]{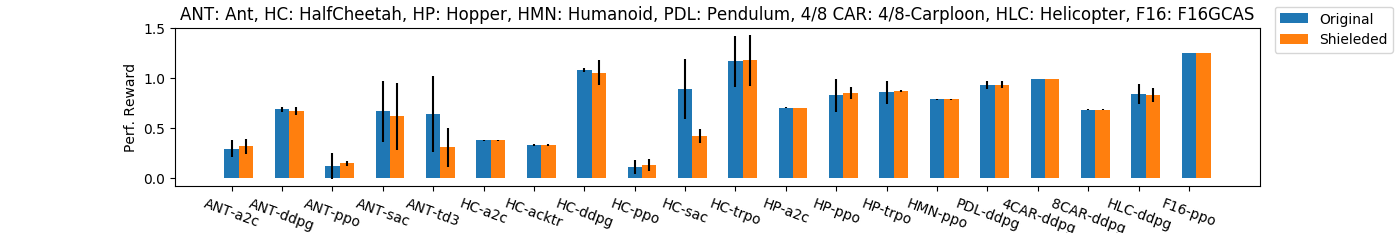}
    \caption{Average Reward for Original Policy and Shielded Policy}
    \label{fig:eval_perf}
\end{figure}
\paragraph{Performance of the Shielded Policy}
\label{sec:perf of shield}
The experiments in Figure \ref{fig:eval_perf} investigate the impact
of considering safety (and thus shield interventions) on performance.
We randomly initialized the simulation around the attacked trajectory
within a range bounded by $\varepsilon$. For each benchmark, we run a
simulation 1000 times and count the (normalized) average return
yielded with the original policy and shielded policy. The y-axis is
the trajectory return scaled by a constant shared among each
environment's benchmarks. It represents the performance of the
corresponding policy(i.e., moving fast with low electricity cost). The result in Figure \ref{fig:eval_perf}
supports our claim that the cost of shield interventions is typically
modest, on average only 5.78\% variance compared to the performance
exhibited by the original.

\section{Related Work}
\label{sec:related}

As discovered in \cite{huang_adversarial_2017}, by adding tiny or even invisible perturbations to inputs at each time-step, neural network RL policies are vulnerable to adversarial attacks, which can lead to abnormal system behavior and significant performance drop. Simple generation algorithms such as fast gradient sign method (FSGM)~\cite{goodfellow_explaining_2015} can craft adversarial examples for some RL algorithms that are less resistant to adversarial attack. However, our approach can effectively attack a wide range of modern RL algorithms. In the black-box setting where gradient information is not available, \cite{behzadan_vulnerability_2017} proposed a policy induction attack that exploits the transferability of adversarial examples~\cite{szegedy_intriguing_2014}. To attack, it obtains adversarial state perturbations from a replica of the victim's network, e.g. using FSGM, and trains the replica to simulate the victim's network. In contrast, our approach does not attempt to replicate the victim's network but uses Bayesian optimization and dimension reduction to approximate only part of the search space that is relevant to insecure model behavior. \cite{lin_tactics_2017} also proposed to find adversarial attacks in the black-box setting. It pushes the RL agent to achieve the expected state under the current state by generating adversarial perturbations from action sequence planning enabled by model-based learning. As opposed to this approach, our work is model-free as we do not need to learn an environment model to guide the search of adversarial examples. While using Bayesian optimization for crafting adversarial examples was previously explored in~\cite{ghosh_verifying_2018}, that work focuses on environment parameters such as initial and goal states. Our work exploits Bayesian optimization to attack at various time-steps in a rollout. In~\cite{GD+20}, the attacker solves an RL problem to generate an adversarial policy creating natural observations and acting in a multi-agent environment to defeat the victim policy. The setting of our problem is substantially different as we assume the input distribution to the victim policy is not significantly changed at test time to better mirror realistic behaviors. Importantly, our technique differs from these previous approaches by training an auxiliary policy to shield and defend the victim network.

To defend an RL policy, prior work has applied adversarial training to improve the robustness of deep RL policies. Adversarial RL training fine-tunes the victim policy against adversary policies by exerting a force vector or randomizing many of the physical properties of the system like friction coefficients~\cite{dexterous,rarl,Mandlekar_2017,Pattanaik_2018}. The trained policy can robustly operate in the presence of adversarial polices that apply disturbance forces to the system. Defenses based on this idea also generalize to multi-agent RL environments~\cite{GD+20} which shows that repeated fine-tuning can provide protection to a victim policy against a range of adversarial opponents. Our approach differs from these techniques because we consider black-box settings, and do not attempt to fine-tune the internals of a victim policy. Instead, a shield-based auxiliary policy is trained to improve the victim policy's resistance to adversarial perturbations. There exists recent work that synthesizes verified shields to protect RL policies leveraging formal methods ~\cite{zhu_inductive_2019,mps,cpsshield}. However, unlike our approach, these techniques do not scale to high-dimensional environment and adversarial disturbances.

\section{Conclusions}
\label{sec:conc}

Learning-enabled controllers for CPS systems are subject to
adversarial attacks in which small perturbations to the states
generated by an environment in response to a controller's actions can
lead to violations of important safety conditions.  In this paper, we
present a new framework that uses directed attack generation to learn
defense policies that accurately differentiate between safe and unsafe
states and an auxiliary shielding policy that aims to recover from an
unsafe state.  Notably, our methodology operates in a completely
black-box setting in which environment dynamics are unknown.
Experimental results demonstrate our approach is highly effective over
a range of realistic sophisticated controllers, with only modest
impact on overall performance.

\section{Broader Impact} 

Machine learning has demonstrated impressive success in a variety of
applications relevant to the core thrust of this paper.  In
particular, autonomous systems such as self-driving cars, UUVs, UAVs,
etc. are examples of the kind of learning-enabled CPS systems whose
design and structure are consistent with the applications considered
here.  These systems are characterized by intractably large state
spaces, only a small fraction of which are explored during training.
Because the space of possible environment actions is so large, it is
unrealistic to expect that the environment models learnt used to train
these controllers represent all possible behaviors likely to be
observed in deployment.  These covariate shifts can affect controller
behavior resulting in violations of important safety constraints.
These violations can have significant negative repercussions in light
of the common use case for these applications which often involves
operating in environments with high levels of human activity.  Our
shielding policy shows significant benefit in improving controller
safety without requiring expensive retraining post-deployment.

\bibliography{neurips_2020.bib}
\include{Appendix}

\end{document}

%% file: Appendix.tex
\newpage
\appendix

\section{Benchmark  Constraints}
We present the initial constraints and safety constraints for all benchmarks in this section. 
\subsection{Classical Control}
The classical control benchmarks and pretrained models come from \cite{zhu_inductive_2019}. Initial and safety constraints are in table \ref{tab: cc_details}. 
\begin{table}[!htp]\centering
\caption{Classical Control Benchmarks}\label{tab: cc_details}
\scriptsize
\begin{tabular}{lccc}\toprule
Benchmarks &Initial Constraints &Safety Constraints \\\midrule
Pendulum &\thead{$-0.3 < x_i < 0.3$, for $0 \leq i \leq 1$} &\thead{$-0.5 < x_i < 0.5$, for $0 \leq i \leq 1$} \\\midrule
4CarPloon &\thead{$-0.1 < x_i < 0.1$, for $0 \leq i \leq 6$} &\thead{$-2 < x_0 < 2$, $-0.5 < x_{1, 3, 5} < 0.5,$ \\ $-0.35 < x_2 < 0.35$, $-1 < x_{4, 6} < 1$} \\\midrule
8CarPloon &\thead{$-0.1 < x_i < 0.1$, for $0 \leq i \leq 14$} &\thead{$-2 < x_0 < 2, -0.5 < x_{1, 3, 5, 7, 9, 11, 13} < 0.5$, \\ $-1 < x_{2, 4, 6, 8, 10, 12, 14} < 1$} \\\midrule
Helicoptor &\thead{$-0.002 < x_i < 0.002$, for $0 \leq i \leq 7$ \\ $-0.0023 < x_i < 0.0023$, for $8 \leq i \leq 27$} &\thead{$-10 < x_{13} < 10$, $-9 < x_{14} < 9$ \\ $-8 < x_i < 8$, for $0 \leq i \leq 27 \land i \neq 13, 14$} \\
\bottomrule
\end{tabular}
\end{table}

\subsection{F16GCAS}
\label{appendix:f16}

The F16GCAS benchmark is modelled with 16 variables. The meaning of
each, along with initial space and constraints are shown in
Table \ref{tab: F16_details}. When the initial space is 0, this
variable is always initialized with 0.
\begin{table}[!htp]\centering
\caption{F16 Ground Collision Avoidance System}\label{tab: F16_details}
\scriptsize
\begin{tabular}{llllll}\toprule
Variables &Meanings &Initial Space &Safety Constraints &Units \\\midrule
$V$ &Airspeed &[491, 545] &[300, 2500] &ft/s \\
$\alpha$ &Angle of attach &[0.00337, 0.00374] &[-0.1745, 0.7854] &rad \\
$\beta$ &Angle of side-slip &0 &[-0.5236, 0.5236] &rad \\
$\phi$ &Roll angle &[0.714, 0.793] &(-inf, inf) &rad \\
$\theta$ &Pitch angle &[-1.269, -1.142] &(-inf, inf) &rad \\
$\psi$ &Yaw angle &[-0.793, -0.714] &(-inf, inf) &rad \\
$P$ &Roll rate &0 &(-inf, inf) &rad/s \\
$Q$ &Pitch rate &0 &(-inf, inf) &rad/s \\
$R$ &Yaw rate &0 &(-inf, inf) &rad/s \\
$p_n$ &Northward displacement &0 &(-inf, inf) &ft \\
$p_e$ &Eastward displacement &0 &(-inf, inf) &ft \\
$h$ &Altitude &[3272, 3636] &[0, 45000] &ft \\
$pow$ &Engine thrust &[8.18, 9.09] &(-inf, inf) &lbf \\
$\int N_{z_e}$ &Integral of down force error &0 &[-3, 15] &g's \\
$\int P_{s_e}$ &Integral of stability roll rate error &0 &[-2500, 2500] &rad \\
$\int (N_y+r)_e$ &Integral of side force \& yaw rate error &0 &(-inf, inf) &mixed \\
\bottomrule
\end{tabular}
\end{table}

\subsection{Pybullet}
\label{appendix:pybullet}
For initial constraints, one may refer to
$\mathtt{robot\_locomotors.py}$ in the pybullet
repo\footnote{\href{https://github.com/bulletphysics/bullet3/blob/ac3dc0eea5298109c3755c45bc497f06f86111f7/examples/pybullet/gym/pybullet_envs/robot_locomotors.py}{pybullet
robot\_locomotors.py}}. The initial spaces of different joints are
defined in the $\mathtt{robot\_specific\_reset()}$ function. 
Safety constraints are defined in the $\mathtt{alive\_bonus()}$
function in the same file.  These constraints focus on 2 variables,
the height $\mathtt{z}$ and the pitch $\mathtt{p}$ of the robot, as well
as the joint contacts with the ground. When the $\mathtt{alive\_bonus()}$
returns a value that is smaller than 0, the agent violates these constraints and the simulation terminates immediately.  We present
the safety constraints in Table \ref{tab: pybullet_constraints}.

\begin{table}[H]\centering
\caption{PyBullet Benchmarks safety constraints}\label{tab: pybullet_constraints}
\scriptsize
\begin{tabular}{lrr}\toprule
Benchmarks &Safety Constraints \\\midrule
Hopper &$\mathtt{z} > 0.8 $ $\land$ $\mathtt{|p|} < 1$ \\
HalfCheetah & \thead{$\neg contact(joint 1, 2, 4, 5) \land |\mathtt{p}| < 1$} \\
Ant &$\mathtt{z} > 0.26$ \\
Humanoid &$\mathtt{z} > 0.78$ \\
\bottomrule
\end{tabular}

\noindent $\neg contact$ means that the joints do not contact with the ground.
\end{table}

\section{Attack Transferability}
\begin{figure}[!h]
    \centering
    \begin{minipage}[!H]{\linewidth}
        \includegraphics[width=0.33\linewidth]{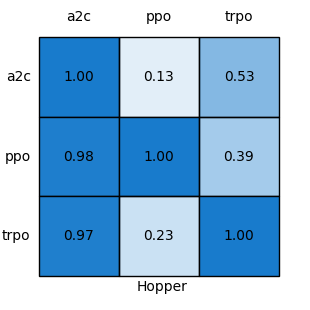}
        \includegraphics[width=0.33\linewidth]{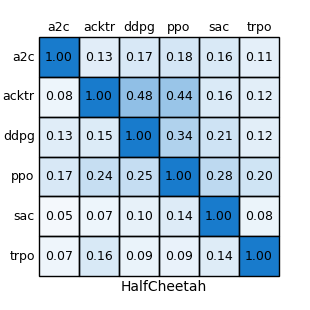}
        \includegraphics[width=0.33\linewidth]{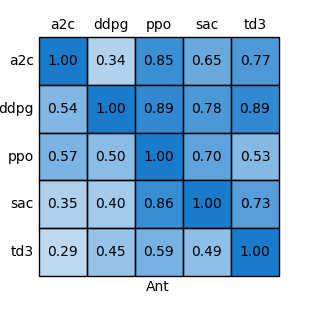}
    \end{minipage}
    \caption{Attack transferability between a given robot's policies. Each row is the adversarial state ratio of one adversarial state set on different policies. i.e, for Hopper, the row a2c, column ppo means $13 \%$ of the adversarial state set of a2c are also adversary states for ppo. The diagonal is always $100 \%$ because states of one policy's adversary dataset are always unsafe for that algorithm.  The numbers in the tables are the transferable ratios.}
    \label{fig:attck_transferability}
\end{figure}

Transferability is a general property of many adversarial attacks.
Recall that given a policy $\pi_0$ in a system $M$ with initial states
$S_0$, an adversarial set $\advset (\pi_0, M, S_0)$ is discovered with
a Bayesian Optimization (BO) attack.  Now, considering another policy
$\pi_1$ on the same system $M$, we wish to know what percentage of
$\advset (\pi_0, M, S_0)$ is also an adversarial state for policy
$\pi_1$. We call this the \emph{transferable ratios} for the benchmark. Several policies trained with different RL algorithms are
available on the Ant, HalfCheetah, and Hopper environments. We measured the adversarial
transferability between these different policies trained for the same
robot system. Figure~\ref{fig:attck_transferability} shows these
results.

Because adversarial states can be transferred between different
policies, the transferable ratio varies on different systems and
policies.  For HalfCheetah, adversarial states are less likely to be
transferred into other policies when compared with Hopper and Ant. It
is also true that an adversary state set can have a relatively high
transferable ratio overall policies. For example, the adversarial
state of Ant-ddpg(row) has greater than a $78\%$ transferable ratio
on all policies. On the contrary, an adversary state set can also have
a low transferable ratio for all policies excepted for its own as
evidenced by HalfCheetah-sac(row). A policy can be vulnerable or
resistant to the adversarial state of other policies. For example,
Hopper-a2c(column) has $98\%$ and $97\%$ transferable ratio on
adversarial states for ppo and trpo respectively, while the
Hopper-ppo(column) has only $13\%$ and $23\%$ transferable ratio on
adversarial states for a2c and trpo, respectively.

\section{Feature Importance}

\begin{figure}[!tp]
    \centering \setlength{\leftskip}{-110pt}
    \begin{minipage}[!H]{1.4\linewidth}
        \includegraphics[scale=0.68]{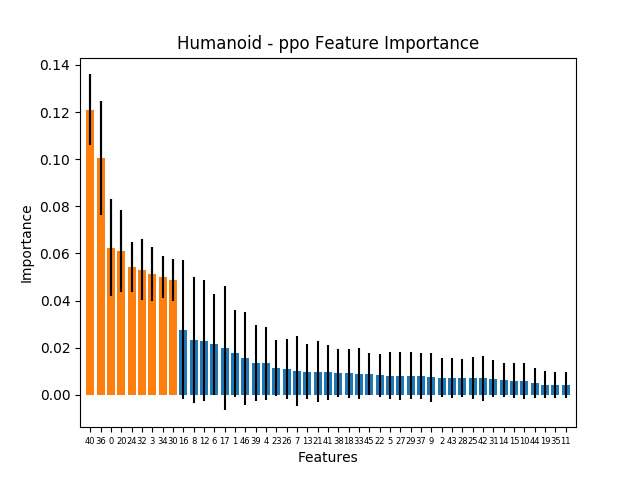}
        \includegraphics[scale=0.68]{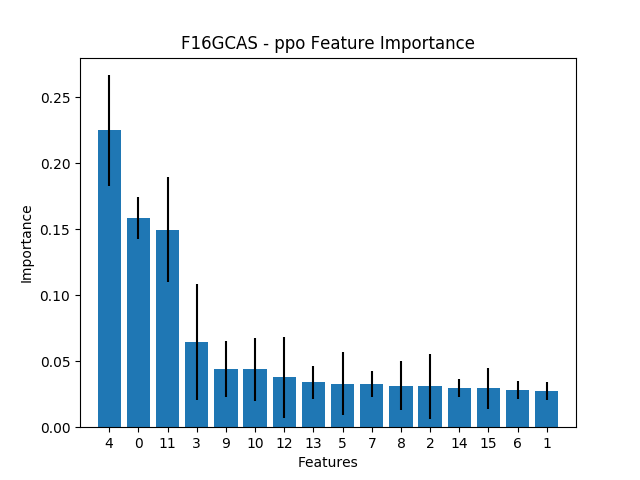}
    \end{minipage}
    \caption{Humanoid-ppo \& F16GCAS-ppo Feature Importance}
    \label{fig:feature_importance}
\end{figure}

Not all features in an application are equally related to safety
specifications. Taking the Humanoid robot as an example - its safety
specification $\varphi$ is to stay upright ($z > 0.78$). This goal is
directly dependent on the height of the robot, thus the height feature
is highly related to safety.  By applying feature selection on the
attack results, we can rank feature importance automatically. In our
experiments, we selected the top-$20\%$ most important features to
reduce the attack dimension (the dimension of filter $\phi$ defined in
Section \ref{sec:approach}) of Humanoid-ppo. Their importance is
marked as orange in Figure~\ref{fig:feature_importance}. We also analyzed the
feature importance of the F16 Ground Collision Avoidance System in
Figure~\ref{fig:feature_importance}. The top-$3$ important features
related to the ground collision are airspeed, pitch angle, and
attitude in order.


\section{Shield Intervention Analysis}

If the shield does not adequately intervene with the original policy,
the original policy is vulnerable to states found in $\advset$.  On
the contrary, since our auxiliary policies are only trained to assist
the original policy to be safe, too many shield interventions
initiated by auxiliary policies can reduce the performance of the
system.  To measure the impact of shield interventions, we crafted
experiments to control the number of interventions. Given an input
state $s$, our detector gives scores to safe and unsafe labels, and
outputs the label with the higher score. For state $s$, suppose the
scores which the detector gives to the safe and unsafe labels are
$S_{safe}(s)$ and $S_{unsafe}(s)$ respectively. In the general
setting, if $S_{safe}(s) > S_{unsafe}(s)$, the detector marks $s$ as
safe.  Now, we change this setting as the detector can mark $s$ safe
iff $S_{safe}(s) > S_{unsafe}(s) + C$. Otherwise, mark $s$ as
unsafe. $C$ is an \emph{approximating constant} - as it increases,
more states will be marked as unsafe, thus there will be more shield
interventions triggered by the auxiliary policy. For small values of
C, more states will be marked as safe, and thus fewer shield
interventions will be triggered.

Thus, by changing the value of the approximating constant, we can measure
the relationship between defense success rates and shield interventions,
as well as the relationship between performance and shield
interventions. For the first case, we initialized our simulation with
all the states in the adversarial state set $\advset(\pi_0, M, S_0)$
for original policy $\pi_0$. In this case, if there is no shield
intervention, the system will violate the safety specification. For
the second one, we measured the performance reward by initializing the
system around one trajectory, which is the same setting as described
in Section \ref{sec:perf of shield}. We decide the range of the
approximating constant with the detector training set derived from the
attack phase.  For each state $s$ in the training set, we calculated
the difference $S_{safe}(s) - S_{unsafe}(s)$ and selected the smallest
and largest difference as the lower bound $l$ and upper bound $h$ of
the approximating constant. Given a sample range $[l, h]$, when $C >
h$, all states in this set will be marked as unsafe; when $C < h$,
all states in this set will be marked as safe. For each benchmark, we
sample 10 approximating constant values in this interval equally
separated.

\begin{figure}[!htp]
    \centering \setlength{\leftskip}{-70pt}
    \begin{minipage}[!H]{\linewidth}
        \includegraphics[scale=0.7]{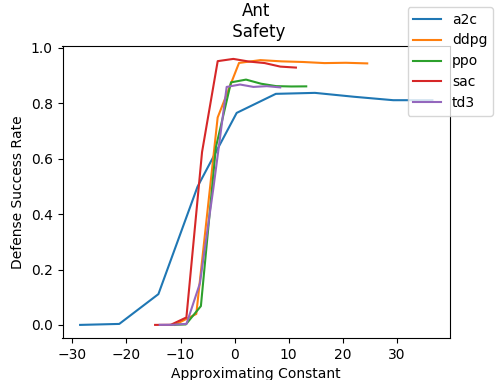}
        \includegraphics[scale=0.7]{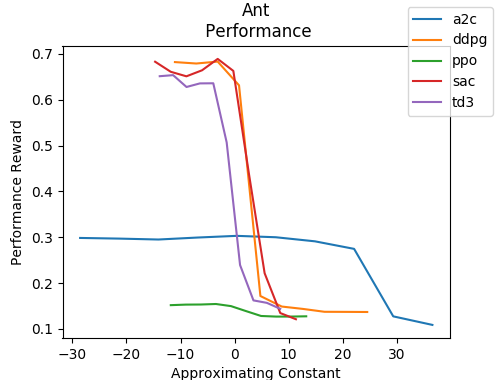}
    \end{minipage}
    \caption{Ant Shield Intervention Analysis}
    \label{fig:ant_shield_intv_analysis}
\end{figure}

\begin{figure}[!htp]
    \centering \setlength{\leftskip}{-70pt}
    \begin{minipage}[!H]{\linewidth}
        \includegraphics[scale=0.7]{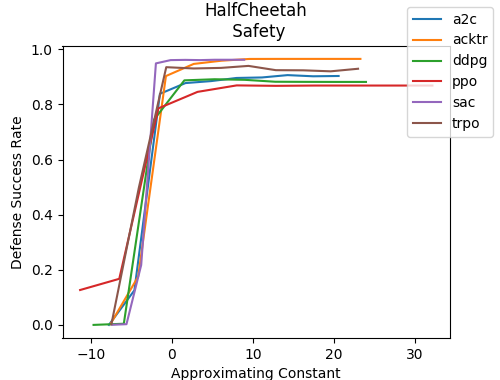}
        \includegraphics[scale=0.7]{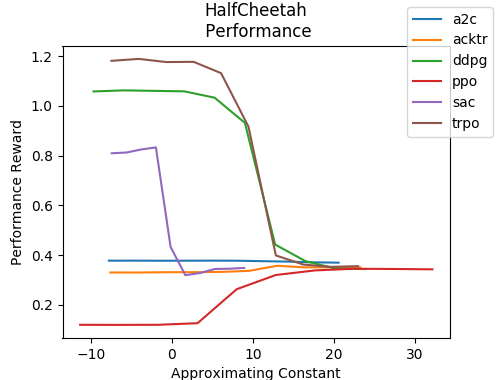}
    \end{minipage}
    \caption{HalfCheetah Shield Intervention Analysis}
    \label{fig:halfcheetah_shield_intv_analysis}
\end{figure}

\begin{figure}[!htp]
    \centering \setlength{\leftskip}{-70pt}
    \begin{minipage}[!H]{\linewidth}
        \includegraphics[scale=0.7]{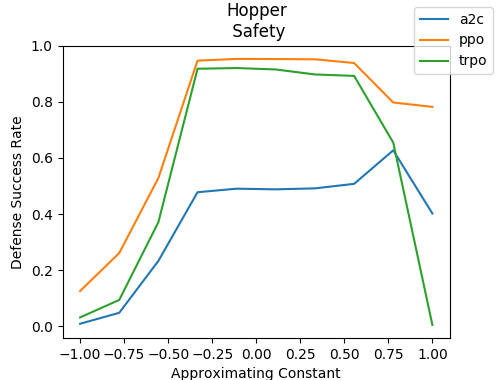}
        \includegraphics[scale=0.7]{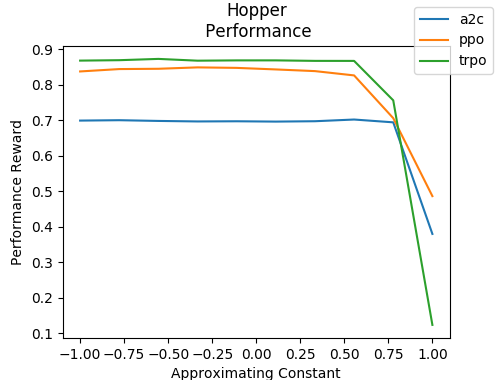}
    \end{minipage}
    \caption{Hopper Shield Intervention Analysis}
    \label{fig:hopper_shield_intv_analysis}
\end{figure}

\begin{figure}[!htp]
    \centering \setlength{\leftskip}{-70pt}
    \begin{minipage}[!H]{\linewidth}
        \includegraphics[scale=0.7]{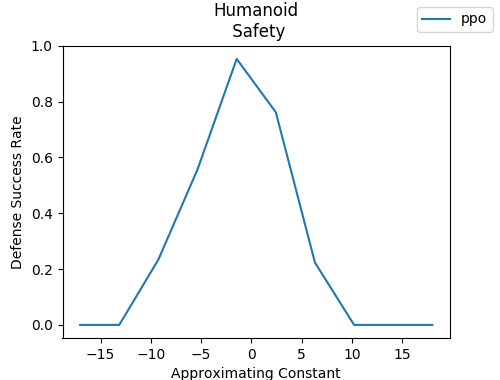}
        \includegraphics[scale=0.7]{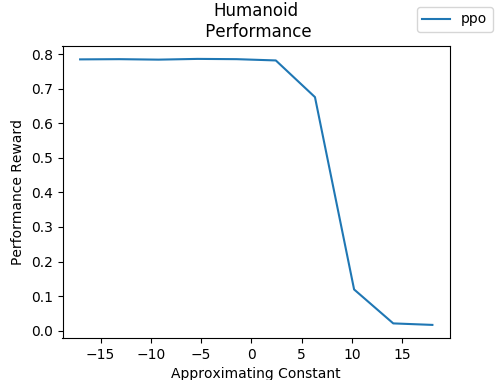}
    \end{minipage}
    \caption{Humanoid Shield Intervention Analysis}
    \label{fig:humanoid_shield_intv_analysis}
\end{figure}

\begin{figure}[!htp]
    \centering \setlength{\leftskip}{-70pt}
    \begin{minipage}[!H]{\linewidth}
        \includegraphics[scale=0.7]{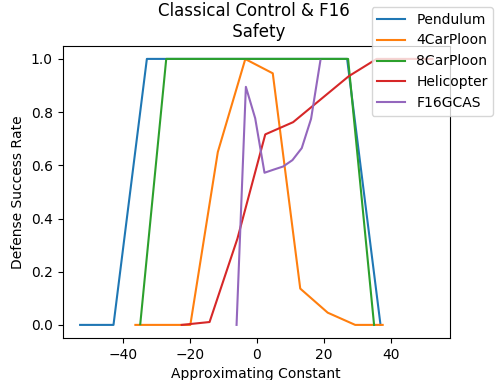}
        \includegraphics[scale=0.7]{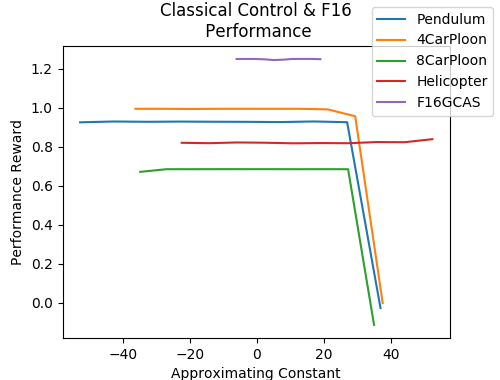}
    \end{minipage}
    \caption{Classical Control \& F16GCAS Shield Intervention Analysis}
    \label{fig:cc_f16_shield_intv_analysis}
\end{figure}

We analyzed shielded policies on Ant in
Fig \ref{fig:ant_shield_intv_analysis}. As the approximating constant
$C$ increases, more interventions are involved and the defense success
rate increases. However, the performance reward decreases dramatically
due to these increasing interventions. The results on HalfCheetah are
shown in Figure~\ref{fig:halfcheetah_shield_intv_analysis}. Similar to
Ant, as $C$ increases, the defense success rate also increases. For
most policies, the performance reward decreases as $C$ increased, but
when policies are not well-trained, such as PPO, the performance
reward actually increases since the system becomes safer and thus the
robot learns a more accurate reward.  The results of Hopper and
Humanoid are shown in Figure~\ref{fig:hopper_shield_intv_analysis} and
Figure~\ref{fig:humanoid_shield_intv_analysis}, respectively.  Unlike
the results shown for Ant and Halfcheetah, we find that too many
interventions can lead to unsafety. This is because the auxiliary
policies are only trained to assist the original policy. As long as it
can pull the system to safe states for the original policies, it can
gain a decent reward. However, the auxiliary policy itself may not be
a safe policy independently.  On the other hand, if the number of
interventions is small, the system is vulnerable to the following
states in $\advset$. Our shielded policies perform best when $C$ is
around 0 since the auxiliary policy is trained under the instruction
of detector when $C$ is $0$.  Similar to Ant and HalfCheetah, when $C
> 0$, the performance reward decreases as the number of interventions
increases on Hopper and Humanoid. Results of classical control and F16GCAS benchmarks are in
Figure \ref{fig:cc_f16_shield_intv_analysis}. Excessive interventions on the (inverted) pendulum, $4$-carploon, and $8$-carploon can make the system unsafe and decrease performance reward.  When interventions are infrequent, these applications become vulnerable to adversarial perturbations. Shielded policies worked best when $C$ is close to 0. For Helicopter and F16GCAS, more interventions lead to safer behavior without comprising performance.

\section{Hyper-parameters}
\subsection{Attack Hyper-parameters}

When running the Bayesian optimization attack, we used the same
hyper-parameters for all benchmarks. The acquisition function is
Expected Improvement (EI). The $\xi$ of EI is 0.01. We use Gaussian
Process (GP) with Matern kernel as a surrogate model. The
hyper-parameters of the Matern kernel use the default tuned values in the
{\tt skopt} \footnote{\href{https://scikit-optimize.github.io/stable/modules/generated/skopt.gp_minimize.html}{skopt.gp\_minimize}}
library.  Before approximating the initial state-reward function with
GP, we randomly sample 10 points.  We then evaluate the initial state-reward
function with points gotten by optimizing on acquisition function 30
times.
\subsection{Defense Hyper-parameters}
The Hopper's detectors are random forests. Each of them has 100 trees
with 50 max depth. The other detectors are neural networks. The search
categories of neural network detectors' hyper-parameters are given in
Table \ref{tab:nn_hyper}.

\begin{table}[H]\centering
    \caption{Hyper-parameters of Neural Network Detector}\label{tab:nn_hyper}
    \scriptsize
    \begin{tabular}{lrr}\toprule
    Hyper-parameter &Search Categories \\\midrule
    Network Structure &obs\_dim x 128 x 128 x 128 x 2 \\
    Learning Rate &[1e-4, 1e-3] \\
    Mini-batch Size &[64, 128, 256, 512, 1024] \\
    Training Epoch &[5, 10, 20, 50] \\
    Optimizer &Adam \\
    \bottomrule
    \end{tabular}
\end{table}

The auxiliary policies of Hopper, HalfCheetah and Ant are trained with DDPG, while the Humanoid, classical control and F16GCAS benchmarks are equipped with PPO auxiliary policy. The search categories of the auxiliary policy are listed in table \ref{tab:ddpg_hyper} and \ref{tab:ppo_hyper}.

\begin{table}[!htp]\centering
\caption{Hyper-parameters of DDPG}\label{tab:ddpg_hyper}
\scriptsize
\begin{tabular}{lrr}\toprule
Parameters & Search Categories \\\midrule
Actor Network Structure &obs\_dim x 64 x 64 x action\_dim \\
Critic Network Structure &obs\_dim x 64 x 64 x 1 \\
Discount Factor $\gamma$ & [0.99, 0.999] \\
Update Rate $\tau$ & [1e-3, 1e-4] \\
others & Default values \footnotemark \\
\bottomrule
\end{tabular}
\end{table}
\footnotetext[5]{\href{https://stable-baselines.readthedocs.io/en/master/modules/ddpg.html}{Stable-Baselines DDPG}}

\begin{table}[!htp]\centering
\caption{Hyper-parameters of PPO}\label{tab:ppo_hyper}
\scriptsize
\begin{tabular}{lrr}\toprule
Parameters & Search Categories \\\midrule
Policy Network Structure &obs\_dim x 128 x 128 x action\_dim \\
Training Step &[1e5, 5e5, 1e6, 2e6, 4e6] \\
Clipping Range &[0.01, 0.05, 0.1, 0.2] \\
Discount Factor \((\gamma)\) &[0.99, 0.999, 1] \\
Entropy Cocfficicnt &0.01 \\
GAE \((\lambda)\) &0.95 \\
Gradient Norm Clipping &0.5 \\
Learning Rate &[2.5e-3, 2.5e-4, 2.5e-5] \\
Number of Actors &[4, 8, 16] \\
Optimizer &Adam \\
Training Epochs per Update &[4, 10] \\
Training Mini-batches per Update & [4, 16, 64, 128]\\
Unroll Length/n-step & [64, 128, 256, 512, 1024]\\
Value Function Coefficient &0.5 \\
\bottomrule
\end{tabular}
\end{table}

%% file: neurips_2020.bbl
\begin{thebibliography}{27}
\providecommand{\natexlab}[1]{#1}
\providecommand{\url}[1]{\texttt{#1}}
\expandafter\ifx\csname urlstyle\endcsname\relax
  \providecommand{\doi}[1]{doi: #1}\else
  \providecommand{\doi}{doi: \begingroup \urlstyle{rm}\Url}\fi

\bibitem[Alshiekh et~al.()Alshiekh, Bloem, Ehlers, Könighofer, Niekum, and
  Topcu]{alshiekh_safe_2017}
Mohammed Alshiekh, Roderick Bloem, Ruediger Ehlers, Bettina Könighofer, Scott
  Niekum, and Ufuk Topcu.
\newblock Safe reinforcement learning via shielding.
\newblock In \emph{{The Thirty-Second AAAI Conference on Artificial
  Intelligence (AAAI-18)}}.
\newblock URL \url{http://arxiv.org/abs/1708.08611}.

\bibitem[Bansal et~al.(2018)Bansal, Pachocki, Sidor, Sutskever, and
  Mordatch]{BP+18}
Trapit Bansal, Jakub Pachocki, Szymon Sidor, Ilya Sutskever, and Igor Mordatch.
\newblock Emergent complexity via multi-agent competition.
\newblock In \emph{6th International Conference on Learning Representations,
  {ICLR} 2018, Vancouver, BC, Canada, April 30 - May 3, 2018, Conference Track
  Proceedings}, 2018.
\newblock URL \url{https://openreview.net/forum?id=Sy0GnUxCb}.

\bibitem[Bastani(2019)]{mps}
Osbert Bastani.
\newblock Safe reinforcement learning via online shielding.
\newblock \emph{CoRR}, abs/1905.10691, 2019.
\newblock URL \url{http://arxiv.org/abs/1905.10691}.

\bibitem[Behzadan and Munir()]{behzadan_vulnerability_2017}
Vahid Behzadan and Arslan Munir.
\newblock Vulnerability of deep reinforcement learning to policy induction
  attacks.
\newblock URL \url{http://arxiv.org/abs/1701.04143}.

\bibitem[Coumans and Bai(2016--2019)]{coumans2019}
Erwin Coumans and Yunfei Bai.
\newblock Pybullet, a python module for physics simulation for games, robotics
  and machine learning.
\newblock \url{http://pybullet.org}, 2016--2019.

\bibitem[Dosovitskiy et~al.(2017)Dosovitskiy, Ros, Codevilla, L{\'{o}}pez, and
  Koltun]{DR+17}
Alexey Dosovitskiy, Germ{\'{a}}n Ros, Felipe Codevilla, Antonio L{\'{o}}pez,
  and Vladlen Koltun.
\newblock {CARLA:} an open urban driving simulator.
\newblock In \emph{1st Annual Conference on Robot Learning, CoRL 2017, Mountain
  View, California, USA, November 13-15, 2017, Proceedings}, pages 1--16, 2017.
\newblock URL \url{http://proceedings.mlr.press/v78/dosovitskiy17a.html}.

\bibitem[Frehse et~al.()Frehse, Le~Guernic, Donzé, Cotton, Ray, Lebeltel,
  Ripado, Girard, Dang, and Maler]{gopalakrishnan_spaceex_2011}
Goran Frehse, Colas Le~Guernic, Alexandre Donzé, Scott Cotton, Rajarshi Ray,
  Olivier Lebeltel, Rodolfo Ripado, Antoine Girard, Thao Dang, and Oded Maler.
\newblock {SpaceEx}: Scalable verification of hybrid systems.
\newblock In Ganesh Gopalakrishnan and Shaz Qadeer, editors, \emph{Computer
  Aided Verification}, volume 6806, pages 379--395. Springer Berlin Heidelberg.
\newblock ISBN 978-3-642-22109-5 978-3-642-22110-1.
\newblock \doi{10.1007/978-3-642-22110-1_30}.
\newblock URL \url{http://link.springer.com/10.1007/978-3-642-22110-1_30}.
\newblock Series Title: Lecture Notes in Computer Science.

\bibitem[Ghosh et~al.()Ghosh, Berkenkamp, Ranade, Qadeer, and
  Kapoor]{ghosh_verifying_2018}
Shromona Ghosh, Felix Berkenkamp, Gireeja Ranade, Shaz Qadeer, and Ashish
  Kapoor.
\newblock {Verifying Controllers Against Adversarial Examples with Bayesian
  Optimization}.
\newblock In \emph{2018 {IEEE} International Conference on Robotics and
  Automation ({ICRA})}, pages 7306--7313.
\newblock \doi{10.1109/ICRA.2018.8460635}.
\newblock URL \url{http://arxiv.org/abs/1802.08678}.

\bibitem[Gleave et~al.(2020)Gleave, Dennis, Wild, Kant, Levine, and
  Russell]{GD+20}
Adam Gleave, Michael Dennis, Cody Wild, Neel Kant, Sergey Levine, and Stuart
  Russell.
\newblock Adversarial policies: Attacking deep reinforcement learning.
\newblock In \emph{8th International Conference on Learning Representations,
  {ICLR} 2020, Addis Ababa, Ethiopia, April 26-30, 2020}, 2020.
\newblock URL \url{https://openreview.net/forum?id=HJgEMpVFwB}.

\bibitem[Goodfellow et~al.()Goodfellow, Shlens, and
  Szegedy]{goodfellow_explaining_2015}
Ian~J. Goodfellow, Jonathon Shlens, and Christian Szegedy.
\newblock Explaining and harnessing adversarial examples.
\newblock URL \url{http://arxiv.org/abs/1412.6572}.

\bibitem[Heidlauf et~al.(2018)Heidlauf, Collins, Bolender, and
  Bak]{heidlauf2018verification}
Peter Heidlauf, Alexander Collins, Michael Bolender, and Stanley Bak.
\newblock Verification challenges in f-16 ground collision avoidance and other
  automated maneuvers.
\newblock In \emph{ARCH@ ADHS}, 2018.

\bibitem[Hill et~al.(2018)Hill, Raffin, Ernestus, Gleave, Kanervisto, Traore,
  Dhariwal, Hesse, Klimov, Nichol, Plappert, Radford, Schulman, Sidor, and
  Wu]{stable-baselines}
Ashley Hill, Antonin Raffin, Maximilian Ernestus, Adam Gleave, Anssi
  Kanervisto, Rene Traore, Prafulla Dhariwal, Christopher Hesse, Oleg Klimov,
  Alex Nichol, Matthias Plappert, Alec Radford, John Schulman, Szymon Sidor,
  and Yuhuai Wu.
\newblock Stable baselines.
\newblock \url{https://github.com/hill-a/stable-baselines}, 2018.

\bibitem[Huang et~al.()Huang, Papernot, Goodfellow, Duan, and
  Abbeel]{huang_adversarial_2017}
Sandy Huang, Nicolas Papernot, Ian Goodfellow, Yan Duan, and Pieter Abbeel.
\newblock Adversarial attacks on neural network policies.
\newblock URL \url{http://arxiv.org/abs/1702.02284}.

\bibitem[Jones et~al.(1998)Jones, Schonlau, and Welch]{jones1998efficient}
Donald~R Jones, Matthias Schonlau, and William~J Welch.
\newblock Efficient global optimization of expensive black-box functions.
\newblock \emph{Journal of Global optimization}, 13\penalty0 (4):\penalty0
  455--492, 1998.

\bibitem[Lin et~al.()Lin, Hong, Liao, Shih, Liu, and Sun]{lin_tactics_2017}
Yen-Chen Lin, Zhang-Wei Hong, Yuan-Hong Liao, Meng-Li Shih, Ming-Yu Liu, and
  Min Sun.
\newblock Tactics of adversarial attack on deep reinforcement learning agents.
\newblock In \emph{Proceedings of the Twenty-Sixth International Joint
  Conference on Artificial Intelligence}, pages 3756--3762. International Joint
  Conferences on Artificial Intelligence Organization.
\newblock ISBN 978-0-9992411-0-3.
\newblock \doi{10.24963/ijcai.2017/525}.
\newblock URL \url{https://www.ijcai.org/proceedings/2017/525}.

\bibitem[Mandlekar et~al.(2017)Mandlekar, Zhu, Garg, Fei{-}Fei, and
  Savarese]{Mandlekar_2017}
Ajay Mandlekar, Yuke Zhu, Animesh Garg, Li~Fei{-}Fei, and Silvio Savarese.
\newblock Adversarially robust policy learning: Active construction of
  physically-plausible perturbations.
\newblock In \emph{2017 {IEEE/RSJ} International Conference on Intelligent
  Robots and Systems, {IROS} 2017, Vancouver, BC, Canada, September 24-28,
  2017}, pages 3932--3939. {IEEE}, 2017.
\newblock \doi{10.1109/IROS.2017.8206245}.
\newblock URL \url{https://doi.org/10.1109/IROS.2017.8206245}.

\bibitem[OpenAI et~al.(2018)OpenAI, Andrychowicz, Baker, Chociej,
  J{\'{o}}zefowicz, McGrew, Pachocki, Pachocki, Petron, Plappert, Powell, Ray,
  Schneider, Sidor, Tobin, Welinder, Weng, and Zaremba]{dexterous}
OpenAI, Marcin Andrychowicz, Bowen Baker, Maciek Chociej, Rafal
  J{\'{o}}zefowicz, Bob McGrew, Jakub~W. Pachocki, Jakub Pachocki, Arthur
  Petron, Matthias Plappert, Glenn Powell, Alex Ray, Jonas Schneider, Szymon
  Sidor, Josh Tobin, Peter Welinder, Lilian Weng, and Wojciech Zaremba.
\newblock Learning dexterous in-hand manipulation.
\newblock \emph{CoRR}, abs/1808.00177, 2018.
\newblock URL \url{http://arxiv.org/abs/1808.00177}.

\bibitem[Pattanaik et~al.(2018)Pattanaik, Tang, Liu, Bommannan, and
  Chowdhary]{Pattanaik_2018}
Anay Pattanaik, Zhenyi Tang, Shuijing Liu, Gautham Bommannan, and Girish
  Chowdhary.
\newblock Robust deep reinforcement learning with adversarial attacks.
\newblock In Elisabeth Andr{\'{e}}, Sven Koenig, Mehdi Dastani, and Gita
  Sukthankar, editors, \emph{Proceedings of the 17th International Conference
  on Autonomous Agents and MultiAgent Systems, {AAMAS} 2018, Stockholm, Sweden,
  July 10-15, 2018}, pages 2040--2042. International Foundation for Autonomous
  Agents and Multiagent Systems Richland, SC, {USA} / {ACM}, 2018.
\newblock URL \url{http://dl.acm.org/citation.cfm?id=3238064}.

\bibitem[Pinto et~al.(2017)Pinto, Davidson, Sukthankar, and Gupta]{rarl}
Lerrel Pinto, James Davidson, Rahul Sukthankar, and Abhinav Gupta.
\newblock Robust adversarial reinforcement learning.
\newblock In Doina Precup and Yee~Whye Teh, editors, \emph{Proceedings of the
  34th International Conference on Machine Learning, {ICML} 2017, Sydney, NSW,
  Australia, 6-11 August 2017}, volume~70 of \emph{Proceedings of Machine
  Learning Research}, pages 2817--2826. {PMLR}, 2017.
\newblock URL \url{http://proceedings.mlr.press/v70/pinto17a.html}.

\bibitem[Rasmussen(2003)]{rasmussen2003gaussian}
Carl~Edward Rasmussen.
\newblock Gaussian processes in machine learning.
\newblock In \emph{Summer School on Machine Learning}, pages 63--71. Springer,
  2003.

\bibitem[Schulman et~al.(2017)Schulman, Wolski, Dhariwal, Radford, and
  Klimov]{DBLP:journals/corr/SchulmanWDRK17}
John Schulman, Filip Wolski, Prafulla Dhariwal, Alec Radford, and Oleg Klimov.
\newblock Proximal policy optimization algorithms.
\newblock \emph{CoRR}, abs/1707.06347, 2017.
\newblock URL \url{http://arxiv.org/abs/1707.06347}.

\bibitem[Sch{\"u}rmann and Althoff(2017)]{schurmann2017optimal}
Bastian Sch{\"u}rmann and Matthias Althoff.
\newblock Optimal control of sets of solutions to formally guarantee
  constraints of disturbed linear systems.
\newblock In \emph{2017 American Control Conference (ACC)}, pages 2522--2529.
  IEEE, 2017.

\bibitem[Silver et~al.(2014)Silver, Lever, Heess, Degris, Wierstra, and
  Riedmiller]{pmlr-v32-silver14}
David Silver, Guy Lever, Nicolas Heess, Thomas Degris, Daan Wierstra, and
  Martin Riedmiller.
\newblock Deterministic policy gradient algorithms.
\newblock In Eric~P. Xing and Tony Jebara, editors, \emph{Proceedings of the
  31st International Conference on Machine Learning}, volume~32 of
  \emph{Proceedings of Machine Learning Research}, pages 387--395, Bejing,
  China, 22--24 Jun 2014. PMLR.
\newblock URL \url{http://proceedings.mlr.press/v32/silver14.html}.

\bibitem[Szegedy et~al.()Szegedy, Zaremba, Sutskever, Bruna, Erhan, Goodfellow,
  and Fergus]{szegedy_intriguing_2014}
Christian Szegedy, Wojciech Zaremba, Ilya Sutskever, Joan Bruna, Dumitru Erhan,
  Ian Goodfellow, and Rob Fergus.
\newblock Intriguing properties of neural networks.
\newblock In \emph{International Conference on Learning Representations}.
\newblock URL \url{http://arxiv.org/abs/1312.6199}.

\bibitem[Wang et~al.(2013)Wang, Zoghi, Hutter, Matheson, and
  De~Freitas]{wang2013bayesian}
Ziyu Wang, Masrour Zoghi, Frank Hutter, David Matheson, and Nando De~Freitas.
\newblock Bayesian optimization in high dimensions via random embeddings.
\newblock In \emph{Twenty-Third International Joint Conference on Artificial
  Intelligence}, 2013.

\bibitem[Wu et~al.(2019)Wu, Wang, Deshmukh, and Wang]{cpsshield}
Meng Wu, Jingbo Wang, Jyotirmoy Deshmukh, and Chao Wang.
\newblock Shield synthesis for real: Enforcing safety in cyber-physical
  systems.
\newblock In \emph{2019 Formal Methods in Computer Aided Design, {FMCAD} 2019,
  San Jose, CA, USA, October 22-25, 2019}, pages 129--137, 2019.
\newblock \doi{10.23919/FMCAD.2019.8894264}.
\newblock URL \url{https://doi.org/10.23919/FMCAD.2019.8894264}.

\bibitem[Zhu et~al.()Zhu, Xiong, Magill, and Jagannathan]{zhu_inductive_2019}
He~Zhu, Zikang Xiong, Stephen Magill, and Suresh Jagannathan.
\newblock An inductive synthesis framework for verifiable reinforcement
  learning.
\newblock In \emph{Proceedings of the 40th {ACM} {SIGPLAN} Conference on
  Programming Language Design and Implementation - {PLDI} 2019}, pages
  686--701. {ACM} Press.
\newblock ISBN 978-1-4503-6712-7.
\newblock \doi{10.1145/3314221.3314638}.
\newblock URL \url{http://dl.acm.org/citation.cfm?doid=3314221.3314638}.

\end{thebibliography}
